\documentclass{elsarticle}

\usepackage{times}
\usepackage{setspace}
\usepackage{graphicx}
\usepackage{epstopdf}
\usepackage{subfigure}
\usepackage{amsmath}
\usepackage{multirow}
\usepackage{algorithm}  
\usepackage{algorithmic}
\usepackage{lineno,hyperref}
\usepackage{color}
\usepackage{amssymb}

\modulolinenumbers[5]
\journal{Signal Processing: Image Communication}
\begin{document}
	
\begin{frontmatter}
\title{A Robust Background Initialization Algorithm with Superpixel Motion Detection}
\author{Zhe Xu}
\ead{zhexu22-c@my.cityu.edu.hk}
\author{Biao Min}
\ead{biaomin3-c@my.cityu.edu.hk}
\author{Ray C.~C. Cheung\corref{correspondingauthor}}
\cortext[correspondingauthor]{Corresponding author}
\ead{r.cheung@cityu.edu.hk}
\address{Department of Electronic Engineering, City University of Hong Kong, Hong Kong, China}

\begin{abstract}
	
	Scene background initialization allows the recovery of a clear image without
	foreground objects from a video sequence, which is generally the first step in
	many computer vision and video processing applications. The process may be
	strongly affected by some challenges such as illumination changes, foreground
	cluttering, intermittent movement, etc. In this paper, a robust background
	initialization approach based on superpixel motion detection is proposed. Both
	spatial and temporal characteristics of frames are adopted to effectively
	eliminate foreground objects. A subsequence with stable illumination condition
	is first selected for background estimation. Images are segmented into
	superpixels to preserve spatial texture information and foreground objects are
	eliminated by superpixel motion filtering process. A low-complexity
	density-based clustering is then performed to generate reliable background
	candidates for final background determination. The approach has been evaluated
	on SBMnet dataset and it achieves a performance superior or comparable to other
	state-of-the-art works with faster processing speed. Moreover, in those complex
	and dynamic categories, the algorithm produces the best results showing the
	robustness against very challenging scenarios.

\end{abstract}

\end{frontmatter}

\section{Introduction}
\label{sec:intro}

Scene background initialization refers to approaches automatically producing a
stationary scene without foreground objects, given a set of training frames. It
is a critical step in many computer vision and video processing applications
such as object detection and tracking~\cite{Panda2016, Zhang2016}, video
segmentation~\cite{Chiu2010}, video coding~\cite{Paul2012, Li2016, Chen2017}, video
inpainting~\cite{Chen2010} and so on. In such applications, obtaining the
background image is generally a pre-processing step and the clear background
can make the processing more effective. 

Several simple background initialization approaches directly choose one frame
from sequence as the background with the assumption that the frame is free of
foreground objects~\cite{BE_Overview}. However this assumption is not feasible
in many video surveillance situations. For example, in crowded shopping malls,
there may exist moving people in every frame so the extracted background is not
accurate. Another simple and intuitive approach is to adopt a temporal median
filter (TMF) for background initialization~\cite{BE_Overview}. TMF can tolerate
moving object noise as long as foreground objects at any position take less than
half of the estimation period. In addition, the computer vision library
OpenCV\footnote{https://opencv.org} provides two prevalent background modeling
methods. One is based on adaptive Gaussian mixture model called
MOG2~\cite{Zivkovic2004} and another is based on k-nearest neighbors
(KNN)~\cite{Zivkovic2006}. 

\begin{figure*}[!t]
	\centering
	\setlength{\leftskip}{0pt plus 1fil minus \marginparwidth}
	\setlength{\rightskip}{\leftskip}
	\includegraphics[width = 5in]{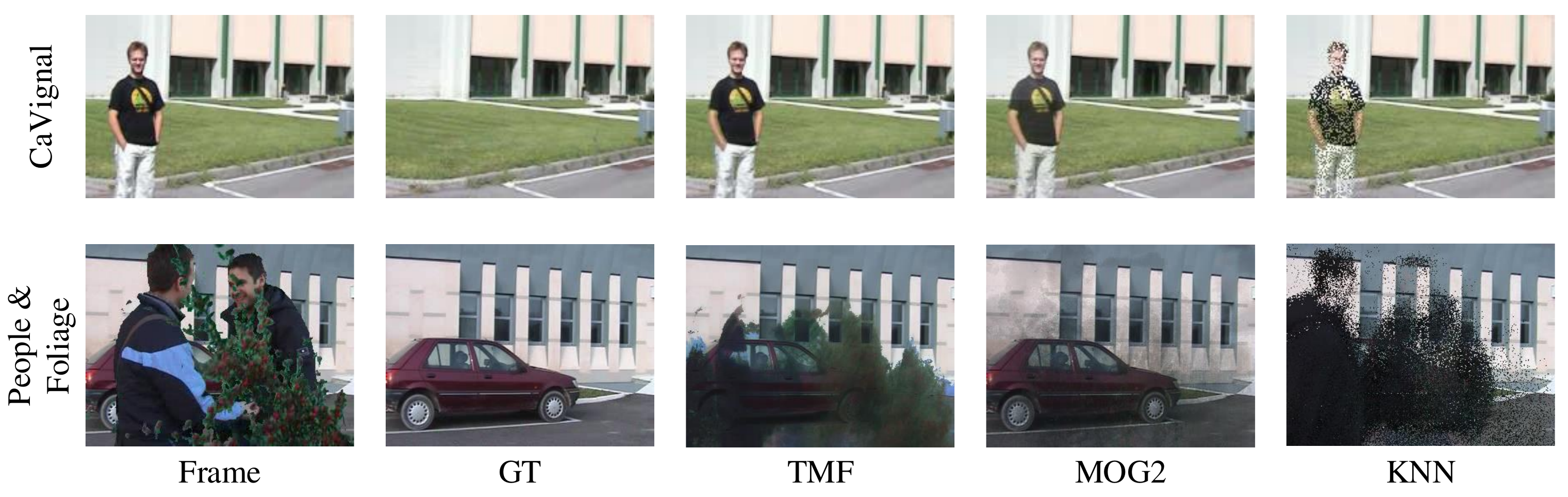}
	\caption{Failing background initialization examples. (The first column
		is one representative frame in original sequence, the second column is the true background
		image and the rest columns are backgrounds obtained by TMF, MOG2 and KNN, respectively.)}
	\label{fig:Drawback}
\end{figure*}

Although the above-mentioned three methods can produce satisfied results
for simple situations, in complex scenarios with heavy foreground cluttering or
long-time stationary objects, most of them cannot eliminate all foreground
components effectively which leads to image blurring. Fig.~\ref{fig:Drawback}
demonstrates two failing background initialization examples. \emph{CaVignal}
shows a situation called intermittent motion, in which the man stands still in
the left of the scene for a long time. All three methods wrongly include the man
in the final estimated background. While in \emph{People \& Foliage}, large
portion of the scene area is occupied by leaves and men. This wide clutter leads
to image blurring especially for TMF and KNN. As a result, a robust background
initialization algorithm is demanded for complex scenes.

In general, the performance of background initialization is affected by
following several factors. Firstly, illumination changes (such as outdoor
sunlight change or indoor lights switching on) may cause illumination variance
in different regions in the estimated background. Secondly, the background in
certain regions may only be visible for a short time so foreground objects are
wrongly included in the result. This occurs when there exists a clutter of
moving foregrounds occupying large portion of the scene area or there exists
stationary foreground object in the estimation period (i.e., intermittent
motion). Thirdly, some motion objects may not be considered as the foreground,
including global motion caused by camera jitter and moving background like
fountain or waving trees. 

In this paper, a robust background initialization algorithm based on superpixel
motion detection with relatively low computational complexity is proposed. Both 
spatial and temporal characteristics of frames are adopted to effectively
eliminate foreground objects. An illumination change detection using histogram
equalization is firstly performed and a subset of sequence with consistent
illumination condition is selected for further processing. Then, simple linear
iterative clustering (SLIC) superpixel algorithm~\cite{Achanta2012} is employed
to segment images to preserve spatial correlations and temporal difference
motion detection is performed in superpixel level to extract motionless regions,
which is called superpixel motion detection (SPMD). A low-complexity
density-based clustering generates reliable background candidates for each pixel
and the final result is selected from candidates with both the first and last
frame as the reference. The motion detection is a superpixel-level strategy
while the density-based clustering is pixel-level. Thus the whole algorithm is a
hybrid method. To extract a clear background in complex scene, some new
approaches are adopted including:

\begin{enumerate}
	\item An illumination change detection scheme is adopted to alleviate
	illumination variance in the final result. Thus for sequences with various
	illumination conditions, the algorithm can produce stable background image. 
	Section~\ref{sec:ICD} will detail this illumination change detection method. 
	\item The superpixel based segmentation is performed to preserve image texture
	information, thus SPMD considers both spatial and temporal correlation. Compared
	with normal pixel-wise motion detection, it can remove relevant moving pixels
    more effectively. The SPMD method is presented in Section~\ref{sec:SPMD}.
\end{enumerate}

The results show that our algorithm can effectively and robustly estimate an
accurate background on SBMnet dataset, which is comparable or superior to
other state-of-the-art works. In addition, the computational complexity is low
making it at least twice faster than other comparable algorithms. The rest of
this paper is organized as follows. Section~\ref{sec:works} reviews previous
related approaches. The proposed robust background initialization algorithm is
introduced in detail in Section~\ref{sec:method}. Experimental results are
provided in Section~\ref{sec:result}. At last, Section~\ref{sec:conclusion}
gives the conclusion.

\section{Related Work}
\label{sec:works}

As a basic step in computer vision and video processing application, background
initialization problem has been extensively studied and corresponding approaches
have been proposed. In this section, some related background initialization
approaches are briefly discussed.

A simple scheme to extract a background is the temporally statistical approach.
In~\cite{Lai1998}, Lai~\emph{et al.} use a running mode and running average
algorithms to initialize the stationary background. A scoreboard is used to keep
the pixel variations. For pixels with large variations, the running mode
algorithm is selected. Otherwise the running average method is used for lower
complexity. In~\cite{Amri2010}, an iterative approach using median blending and
spatial segmentation is proposed. In each iteration, moving objects are detected
based on motion compensation and removed for background initialization. The
median value of the remaining parts is taken as the background.
In~\cite{Wang2006}, a two-step framework is proposed for background
initialization. All stable subsequences are firstly identified based on a frame
difference method. Then the most reliable subsequence is determined and the mean
value of selected subsequence is taken as the background value.
In~\cite{Stauffer1999}, Stauffer~\emph{et al.} propose a background
initialization method by modeling each pixel as a Gaussian Mixture Model (GMM).
Each pixel is classified as background or foreground based on whether the pixel
value fits the current background distribution. In~\cite{Zivkovic2004}, Zivkovic
proposes a similar GMM method for background initialization. The number of
components for each pixel is updated in an online process, thus the algorithm
can automatically adapt to the scene. In~\cite{Laugraud2017}, a background
subtraction algorithm is firstly performed to reduce the set of value candidates
for each pixel. This candidates selection is performed on a patch. Then for each
pixel, the temporal median filter for the candidates is applied to generate a
stable background. 

To avoid parameter tuning problem, some methods use a nonparametric scheme to
initialize the background image. Liu~\emph{et al.}~\cite{Liu2007} present a
background initialization method based on nonparametric model. The most reliable
background mode is calculated based on mean shift clustering and the value is
taken as the estimated background. Elgammal~\emph{et al.}~\cite{Elgammal2000}
introduce a nonparametric background modeling by estimating the probability of pixel
intensity values based on kernel estimator. The pixel is then considered whether
to be a background based on estimated probability. In~\cite{Zhang2012},
Zhang~\emph{et al.} propose a two-stage background initialization method. The
first stage monitors pixel intensity to identify background variations and
creates a look-up table as the intensity distribution. Then in the second stage,
based on whether current pixel is in the look-up table, the final background is
determined.

In recent years, new approaches have been proposed to generate clear background 
images in complex scenes. These algorithms adopt some new techniques such as
matrix or tensor completion, neural networks and so on. In~\cite{Sobral2017},
the background estimation is modeled as a matrix or tensor completion task. The
redundant frames are eliminated first and moving regions are represented by
zeros. Then various matrix or tensor completion methods are utilized to
reconstruct the whole frame sequence with moving objects removed. The background
image is finally initialized as the mean value of the completed sequence.
In~\cite{Gregorio2017}, Gregorio~\emph{et al.} propose a background
initialization approach by modeling video background as ever-changing states of
weightless neural networks. The background estimation of each pixel is produced
by weightless neural networks designed to learn pixel color frequency.
In~\cite{Halfaoui2016}, convolutional neural network (CNN) is employed to
generate background patches from video sequence. A contractive stage with
convolutional layers is performed to extract high-level image features. Then the
refinement stage performs the deconvolution operation to transform the feature
map to the final background patch. In~\cite{WangPP}, Wang~\emph{et al.} present
a joint Gaussian conditional random field (JGCRF) background initialization
algorithm. The target is to find optimal weights for frame fusion and it is
solved as an optimization process with maximum a posteriori problem.
In~\cite{Javed2017}, the background extraction is modeled as a matrix
decomposition problem. The spatial and temporal sparse subspace clustering is
incorporated into robust principal component analysis (RPCA). The low-rank
component is then taken as the estimated background. These recently proposed
algorithms improve the background performance in complex scenes, however the
computational complexity also increases. Thus a robust background initialization
algorithm with low complexity is still needed.

\section{Proposed Background Initialization Algorithm}
\label{sec:method}

Since background initialization process may be heavily affected by some
challenges such as illumination changes, foreground cluttering, intermittent
movement and so on, a robust algorithm is needed. In this section the proposed
background initialization algorithm with low complexity is introduced comprising
four steps. 

\subsection{Overview of the Proposed Approach}

\begin{figure}[!t]
	\centering
	\includegraphics[width = 4in]{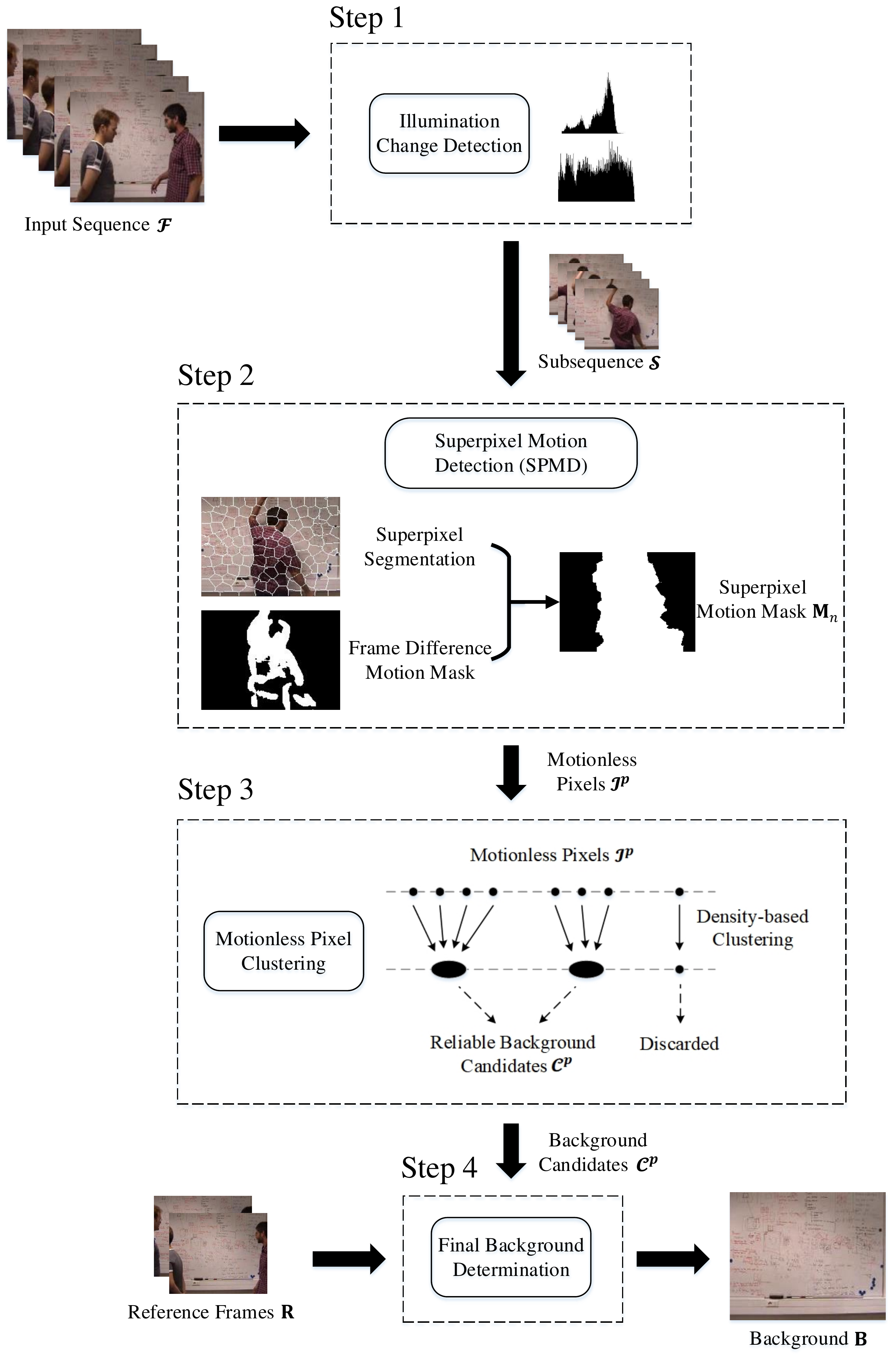}
	\caption{Overview of the proposed background initialization algorithm.}
	\label{fig:Scheme}
\end{figure}

As the background initialization is to eliminate moving objects, motion
detection method is commonly used~\cite{Amri2010,
	Laugraud2017, WangPP, Javed2017, Laugraud2017_2}. One simple approach is the
temporal difference method in which the value differences of pixels between two
consecutive frames are used to detect moving regions. This method is simple to
implement but it does not consider image texture so it may leave holes in the
center of a foreground object when it has uniform texture. Some other
works~\cite{Javed2017, Laugraud2017_2} compute dense optical flow to get motion
information but the computational complexity is high. In this paper, to
effectively detect motion regions with low complexity, the temporal difference
method is combined with image superpixel segmentation called superpixel motion
detection (SPMD). This approach considers both image spatial and temporal
correlations to improve the background initialization performance.

The proposed algorithm comprises the following four steps, which are shown in
Fig.~\ref{fig:Scheme}.
\begin{enumerate}
	\item The input sequence is denoted as $\mathcal{F}=\{\mathbf{F}_t\}_{t=1\dots
		T}$, where $T$ is the length of the sequence and $\mathbf{F}_t$ stands for each
	frame. To cope with the gradual or sudden illumination changes in sequence, the
	illumination change detection by comparing the Hellinger distance of equalized
	histograms is performed. The longest subsequence with stable illumination
	condition is selected denoted as $\mathcal{S}=\{\mathbf{S}_n\}_{n=1\dots N, N
		\leq T}$.
	\item For each frame, the SLIC superpixel segmentation and frame difference
	motion detection are performed called SPMD to generate motion mask
	$\mathbf{M}_n$, where $n$ is the frame index. The superpixel with motion pixels
	in it is classified as foreground and is removed. After SPMD, the selected
	motionless pixels at position $p$ are represented as
	$\mathcal{I}^p=\{i_u^p\}_{u=1\dots U^p}$, where $U^p$ is the number of
	motionless pixels and $U^p\leq N,\forall p$.
	\item Motionless pixels $\mathcal{I}^p$ are then clustered using a
	density-based clustering algorithm. Median value of each cluster is calculated
	as the reliable background candidates $\mathcal{C}^p=\{c_v^p\}_{v=1\dots V^p}$,
	where $V^p$ is the number of cluster at position $p$, and $V^p$ is typically a
	small number.
	\item The final stationary background $\mathbf{B}$ is generated from candidates
	$\mathcal{C}^p$ considering both pixel number in each cluster and distance to
	the corresponding pixel in the first and last frame. 
\end{enumerate}

The proposed algorithm can generate stationary backgrounds for different scenes
robustly. These four steps will be introduced in detail in the following sections.

\subsection{Illumination Change Detection}
\label{sec:ICD}

Illumination change is common in video surveillance due to light switching,
sunlight altering and so on. The main challenge when encountering
illumination change is that light portion and dark portion may both exist in the
result and a smudged background is produced. Thus the illumination change detection 
is significative and it can be included in other scene change detection
applications.

In this section we mainly discuss global illumination changes. The problem is to
distinguish illumination change from pixel intensity change caused by object
movement. HSV color space is utilized because it is more stable than RGB space
when considering illumination condition. HSV space divides the color into three
components: hue (H), saturation (S) and value (V) and the value component is
related to illumination strength. Therefore we employ the histogram and equalized
histogram of V component to analyze the difference. 

\begin{figure}[!t]
	\centering
	\subfigure[Foliage]{\includegraphics[height=2.5in]{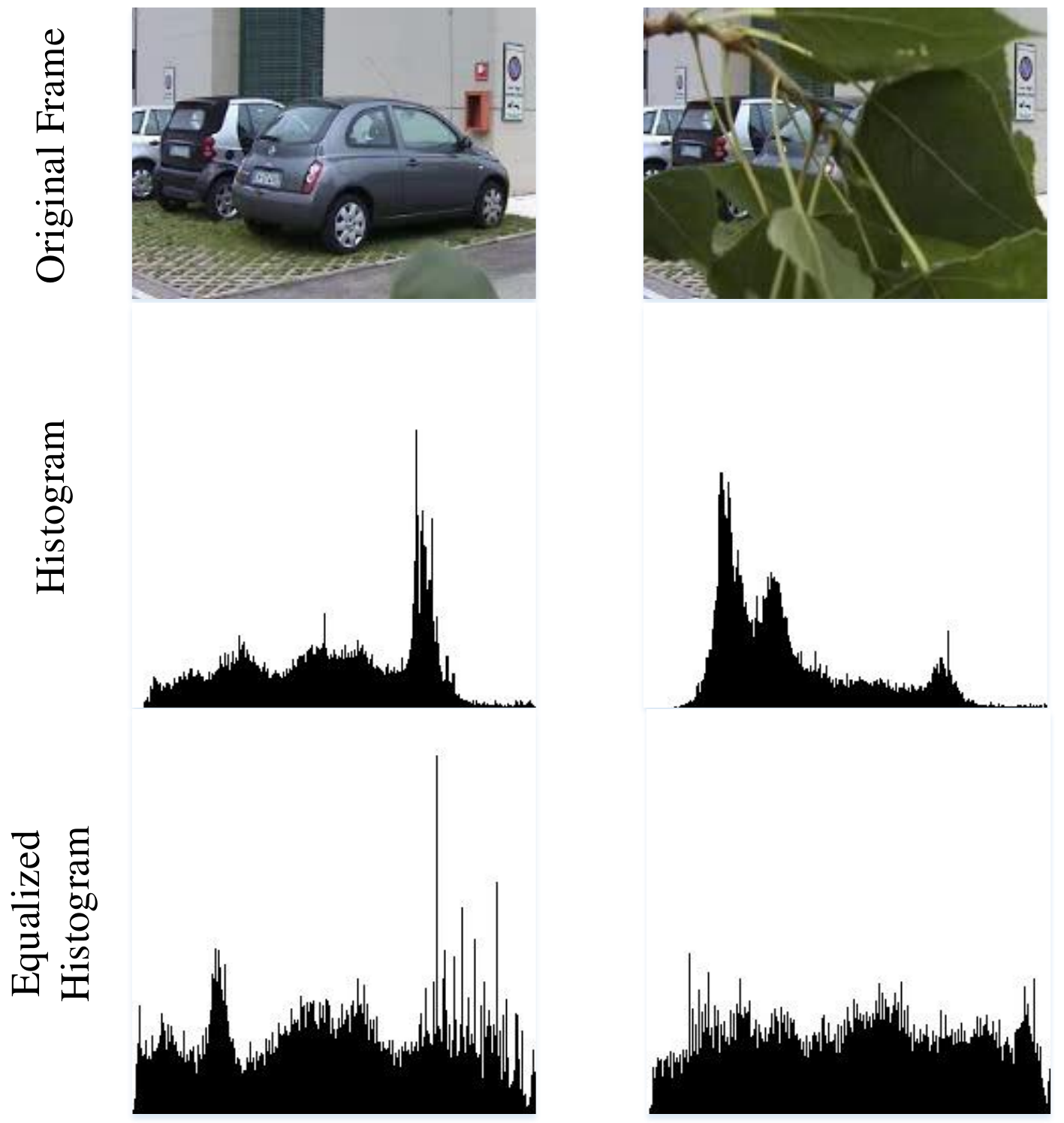}\label{fig:Foliage}}
	\subfigure[Dataset3Camera1]{\includegraphics[height=2.5in]{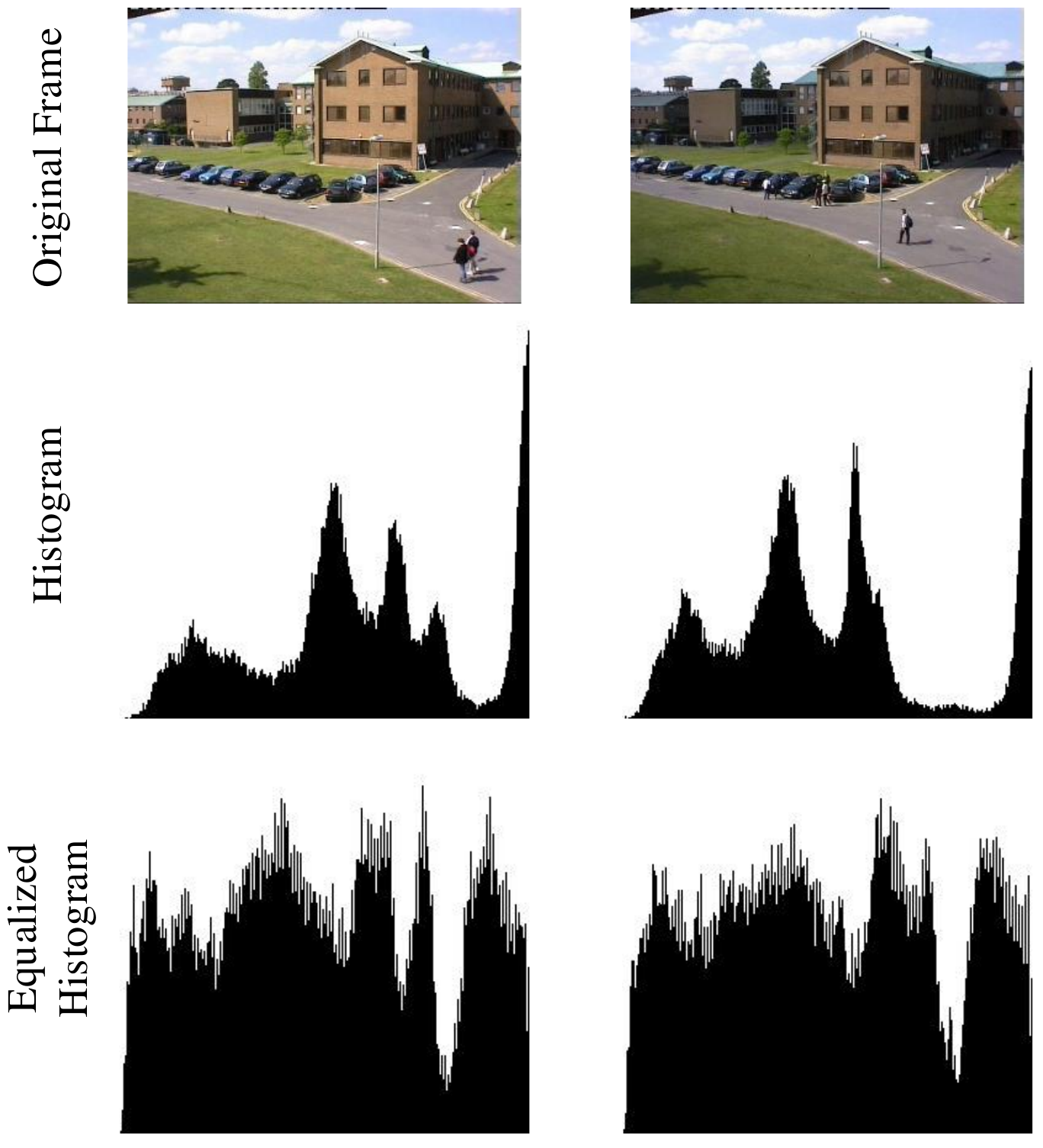}\label{fig:D3C1}}
	\caption{Differences of histogram and equalized histogram caused by object movement and illumination change.}
	\label{fig:Illumination}
\end{figure}

Fig.~\ref{fig:Illumination} shows the histogram and equalized histogram of two
sequences \emph{Foliage} and \emph{Dataset3Camera1}. First row is the original
frame, second row is histogram of V value and third row is the equalized
histogram. In ~\emph{Foliage}, there is no evident illumination change and the
difference of two histogram is caused by movement of leaves. The shapes of two
histograms are very different and even for equalized histogram, the difference
is obvious. For \emph{Dataset3Camera1}, the pixel intensity variance in two
frames is mainly caused by illumination change. The shapes of two histograms are
quite similar only the position of peaks are different. We can observe that
when two histograms are equalized, the difference between them are significantly
reduced. Based on former analysis, if the pixel intensity change is mainly
caused by illumination change, the difference of value component histograms is
obvious, but the equalized histograms are similar.

In the proposed illumination change detection, for a given sequence
$\mathcal{F}$, the first frame is set as the initial reference frame. The
histogram and equalized histogram of current frame are compared with the
reference frame using Hellinger distance $d_{h}$. $d_{h}$ is calculated as:
\begin{equation}
\begin{split}
	d_{h}&=\sqrt{1-bc(H_1,H_2)} \\
	bc(H_1,H_2)&=\frac{\sum_i\sqrt{H_1(i)\cdot H_2(i)}}{\sqrt{\sum_i H_1(i) \cdot \sum_i H_2(i)}}
\end{split}
\end{equation}
$H_1$ and $H_2$ stand for two histograms and $i$ is the histogram bin index.
$bc(H_1,H_2)$ is called Bhattacharyya coefficient of two histograms. Small
$d_{h}$ value means two histograms are similar. We denote $d_{h}(H)$ and
$d_{h}(EH)$ as the Hellinger distance of two histograms and two equalized
histograms, respectively. If there exists illumination change, then following
condition should be satisfied:
\begin{equation}
\begin{cases}
&d_{h}(H) > \tau_{h}    \\
&d_{h}(EH) < \tau_{eh}
\end{cases}
\end{equation}
$\tau_{h}$ and $\tau_{eh}$ are two thresholds controlling illumination detection. 
If $\tau_h$ is too small, the detection is sensitive and the result may be highly
affected by noises. If $\tau_{eh}$ is too large, then object movements may be
wrongly classified. If illumination change is detected, a new subsequence
begins and the reference frame is updated by current frame. After all frames are
tested, the longest subsequence is selected denoted as $\mathcal{S}$ for further
processing.

\subsection{Superpixel Motion Detection (SPMD)}
\label{sec:SPMD}

In SPMD, each frame in $\mathcal{S}$ is segmented using superpixel algorithm.
SLIC algorithm~\cite{Achanta2012} is selected because of its low computational
complexity and high memory efficiency. SLIC algorithm employs the adaptive
k-means clustering to group pixels in CIELAB color space. The clustering
procedure begins with the initial clustering centers $[l_i, a_i, b_i, x_i,
y_i]^T$ with size $L\times L$ in a regular grid. The first three components are
color intensity and the last two components stand for pixel position. $L$ is the
expected approximate superpixel size. Then each pixel is assigned to the nearest
cluster center with limited searching space to reduce complexity. After the
assignment step, the cluster centers are updated to be the mean value of all
pixels belonging to the cluster. The assignment step and update step are
repeated until convergence.

Since SLIC considers a pixel as a 5-dimensional data $[l_i, a_i, b_i, x_i, y_i]^T$
with both color intensity components and spatial components. The distance
measure $d_{sp}$ is altered accordingly and is calculated as 
\begin{equation}
\begin{split}
	d_c&=\sqrt{(l_j-l_i)^2+(a_j-a_i)^2+(b_j-b_i)^2}\\
	d_s&=\sqrt{(x_j-x_i)^2+(y_j-y_i)^2} \\
	d_{sp}&=\sqrt{d_c^2+\biggl(\frac{d_s}{L}\biggr)^2 \cdot m^2} 
\end{split}
\end{equation}
where $d_c$ and $d_s$ are measurements of color proximity and spatial proximity,
respectively. $L$ is the approximate superpixel size and $m$ is a constant. $L$
and $m$ are adopted to normalize color proximity and spatial proximity.

In the proposed algorithm, the superpixel size $L$ is assigned to be adaptive to
image size which is calculated as
\begin{equation}
	L=\biggl\lfloor\frac{min(W, H)}{\sigma_n}\biggr\rfloor
\end{equation}
where $W$ and $H$ are width and height of the image, respectively. $\sigma_n$ is 
the parameter controlling superpixel size. 

\begin{figure}[!t]
	\centering
	\includegraphics[width = 4.5in]{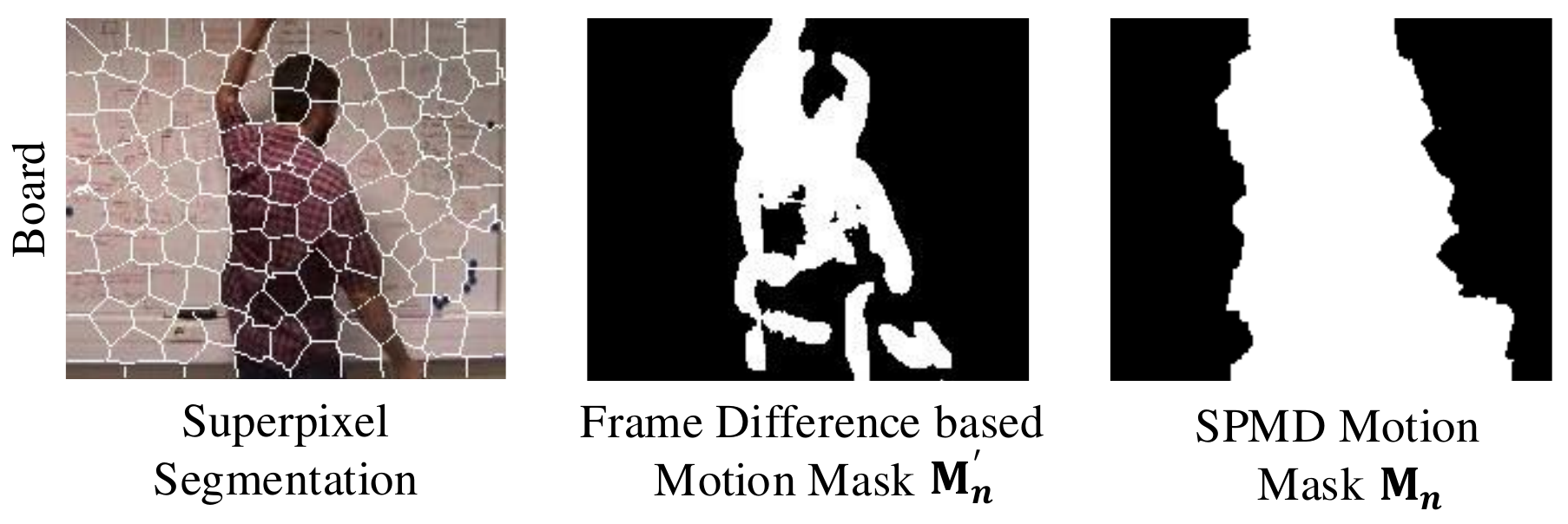}
	\caption{Improvements in motion objects elimination of SPMD.}
	\label{fig:SPMD}
\end{figure}

Besides the superpixel segmentation, for a given frame $\mathbf{S}_n$ in
$\mathcal{S}$, temporal frame difference is calculated as
\begin{equation}
	\mathbf{D}'_n=|\mathbf{S}_n-\mathbf{S}_{n-1}|,n\geq 2
\end{equation}
In our approach, $\mathbf{S}_n$ and $\mathbf{S}_{n-1}$ are converted to the gray
scale in advance so $\mathbf{D}'_n$ is a gray image as well. To avoid some noisy
pixels, Gaussian filtering is performed for each pixel to get the stable
temporal difference $\mathbf{D}_n$. We use Otsu method~\cite{Otsu1979} to
automatically calculate the frame difference based motion mask $\mathbf{M}'_n$
based on $\mathbf{D}_n$. The Otsu method calculates the optimal decision
threshold $\tau_{opt}$ which minimize the intra-class variance, represented as
\begin{equation}
\tau_{opt} = \arg\min_{g}(\omega_0(g)\sigma_0^2(g)+\omega_1(g)\sigma_1^2(g))
\end{equation}
$\omega_0(g)$ and $\omega_1(g)$ are the probabilities of the moving and
motionless pixels when the decision threshold is $g$. $\sigma_0^2(g)$ and
$\sigma_1^2(g)$ are corresponding intra-class variance. Then the frame
difference based motion mask $\mathbf{M}'_n$ can be obtained by
\begin{equation}
m'_n(x,y)=\begin{cases}
1, &d_n(x,y) \geq \tau_{opt}\\
0, &d_n(x,y) < \tau_{opt}
\end{cases}
\end{equation}
where $m'_n(x,y)$ and $d_n(x,y)$ are entries of $\mathbf{M}'_n$ and
$\mathbf{D}_n$ at position $(x,y)$, respectively. Then the final SPMD motion
mask $\mathbf{M}_n$ is determined combining both superpixel segmentation and
$\mathbf{M}'_n$. For each superpixel region $SR_i$, as long as there exist
moving pixels in it, the entire region $SR_i$ is classified as a moving patch.
The calculation of $\mathbf{M}_n$ can be represented as 
\begin{equation}
\begin{split}
&\forall (x,y) \in SR_i: \\
&m_n(x,y)=\begin{cases}
1, &\exists (x',y')\in SR_i: m'_n(x',y')=1\\
0, &otherwise
\end{cases}
\end{split}
\end{equation}

Fig.~\ref{fig:SPMD} shows an example of SPMD. For frame difference based motion
detection, the corresponding motion mask $\mathbf{M}'_n$ leaves many holes in
the center of foreground person because of the uniform texture. By grouping
neighboring pixels in the same object using SLIC, these holes can be correctly
classified as a motion patch. As a result, the unexpected errors caused by
stationary moving objects or objects clustering are reduced.

\subsection{Motionless Pixel Clustering}

After SPMD, motion patches are eliminated and remaining motionless pixels are
denoted as $\mathcal{I}^p=\{i_u^p\}_{u=1\dots U^p}$. $p$ stands for pixel
position and $U^p$ is the number of pixels without motion at position $p$. A
clustering stage is necessary to generate representative pixel intensity as
reliable background candidates $\mathcal{C}^p$ for each position. Density-based
clustering~\cite{Ester1996} is a simple yet efficient clustering algorithm.
Different from other algorithms like k-mean clustering or mixture-of-Gaussian
(MOG), it requires little knowledge of the input data and does not need to set
optimal number of clusters in advance. In the proposed algorithm, the
density-based clustering is employed to generate background candidates. The
pixel intensity is firstly converted to the gray scale, thus the clustering 
process copes with scalar data.

\begin{algorithm}[!t]  
	\caption{Density-based clustering for background candidates generation}  
	\label{alg:Clustering}  
	\begin{algorithmic}  
		\STATE {Input: $\mathcal{I}^p=\{i_u^p\}_{u=1\dots U^p}$} 
		\STATE {Parameter: $\epsilon$, $MinPts$}
		\STATE {Output: 
			\begin{minipage}[t]{0.9\linewidth}
				background candidates $\mathcal{C}^p=\{c_v^p\}_{v=1\dots V^p}$  \\
				pixel number of each cluster $\mathcal{Q}^p=\{q_v^p\}_{v=1\dots V^p}$
			\end{minipage}
		} 
		\STATE {Prerequisite: $\mathcal{I}^p$ have been sorted based on gray intensity} 
		\STATE {Core set $\Omega=\varnothing$}
		
		\FOR {$u = 1,2,\dots, U^p$}
		\STATE {Get $\epsilon$-neighborhood of $i_u^p$: $N_{\epsilon}(i_u^p)$}
		\STATE {Get number of pixels in $N_{\epsilon}(i_u^p)$: $|N_{\epsilon}(i_u^p)|$}
		\IF {$|N_{\epsilon}(i_u^p)|\geq MinPts$}
		\STATE {$i_u^p$ is a core object and add $i_u^p$ into $\Omega$}
		\ENDIF
		\ENDFOR 
		
		\STATE {Number of clustering $V^p=0$} 
		
		\WHILE {$\Omega\neq \varnothing$}
		\STATE {Find smallest pixel in $\Omega$: $i_k^p$}
		\STATE {Find smallest pixel in $N_{\epsilon}(i_k^p)$ as the left boundary of $i_k^p$: $i_l^p$}
		\STATE {Find largest pixel in $N_{\epsilon}(i_k^p)$ as the right boundary of $i_k^p$: $i_r^p$}
		\STATE {Find largest pixel $i_m^p$ in $\Omega$, $s. t.\ i_m^p \leq i_r^p$}
		\WHILE {$i_m^p \neq i_k^p$}
		\STATE {$i_k^p=i_m^p$}
		\STATE {Update $i_r^p$, $i_m^p$}
		\ENDWHILE
		\STATE {$V^p=V^p+1$, Generate a new cluster $\Gamma_v^p=[i_l^p, i_r^p]$}
		\STATE {Remove all core objects in $\Gamma_v^p$ from $\Omega$}
		\STATE {Calculate $c_v^p=median(\Gamma_v^p)$}
		\STATE {Calculate $q_v^p$ as the pixel number in cluster $\Gamma_v$}
		\ENDWHILE
	\end{algorithmic}  
\end{algorithm}

The algorithm relies on a density-based notion of clusters. Four corresponding
concepts defining density-based clustering need to be clarified:
\begin{enumerate}
	\item $\epsilon$-neighborhood of pixel $i_u^p$, denoted as
	$N_{\epsilon}(i_u^p)$, is a pixel set comprising all pixels with distance to
	$i_u^p$ no larger than $\epsilon$. That is, $N_{\epsilon}(i_u^p)=\{i_k^p\ |\
	|i_k^p-i_u^p|\leq\epsilon\}$.
	\item A pixel $i_u^p$ is a core object if its $\epsilon$-neighborhood contains
	at least $MinPts$ pixels, that is, $|N_{\epsilon}(i_u^p)|\geq MinPts$.
	\item A point $i_k^p$ is directly density-reachable from the core object
	$i_u^p$ if $i_k^p$ is in the the $\epsilon$-neighborhood of $i_u^p$.
	\item A pixel $i_k^p$ is density-reachable from the core object $i_u^p$ if
	there exist a chain in $\mathcal{I}^p$: $\{a_1, a_2, \dots, a_n\}, a_1=i_u^p,
	a_n=I_k^p$ such that $a_{i+1}$ is directly density-reachable from $a_i$.
\end{enumerate}

The objective of our density-based clustering is to find all pixels that are
density-reachable from the same core object and group them together to generate a
cluster set $\{\Gamma_v^p\}_{v=1\dots V^p}$ where $V^p$ is the number of cluster
at position $p$. The procedure of the algorithm is shown in
Algorithm~\ref{alg:Clustering}. Each pixel is classified whether to be a core
object or not first. From the smallest core object $i_k^p$, the maximum
density-reachable set is determined to form a cluster $\Gamma_v^p$. Since the
motionless pixels $\mathcal{I}^p$ have been sorted based on gray intensity in
advance, it is easy to find the largest core object $i_m^p$ in
$N_{\epsilon}(i_k^p)$ and this searching is iterated until $i_m^p=i_k^p$. Then, 
the cluster generation starts from a new core object. This iteration continues
until all core objects are classified into a cluster. Each core object is
searched at most one time so the computational complexity is low. At last, the
median filter is applied on each cluster to calculate the background candidates
set $\mathcal{C}^p=\{c_v^p\}_{v=1\dots V^p}$. The pixel number of each cluster
is also recorded and denoted as $\mathcal{Q}^p=\{q_v^p\}_{v=1\dots V^p}$ for
final background determination.

\subsection{Final Background Determination}

The final step is to generate background $\mathbf{B}$ from multiple candidates
$\mathcal{C}^p$. To estimate an accurate background, the decision criteria
considers both pixel number in each cluster $\mathcal{Q}^p$ and original frames
in $\mathcal{S}$ as reference. It is based on the following assumptions:
\begin{enumerate}
	\item SPMD already eliminates moving objects effectively, therefore, background
	component should be dominant in $\mathcal{C}^p$. The candidate with large
	$q_v^p$ value is more likely to be the correct background.
	\item Background has large likelihood to appear in the first frame
	$\mathbf{S}_1$ or the last frame $\mathbf{S}_N$. The mean value of
	$\mathbf{S}_1$ and $\mathbf{S}_N$ is adopted as a reference for decision,
	denoted as $\mathbf{R}$. If a candidate value $c_v^p$ is close to $\mathbf{R}$,
	then it is potential to be selected. It is import to note that SPMD has removed
	most moving objects so even $\mathbf{R}$ includes foreground components, the
	final result is still reliable.
\end{enumerate}

The decision is to choose candidates $c_v^p$ with large $q_v^p$ value and short
distance with reference $\mathbf{R}$. It is represented as 
\begin{equation}
	b^p = \arg\max_{c_v^p} \biggl\{\frac{q_v^p}{|c_v^p-r^p|}\biggr\}
\end{equation}
where $b^p$ and $r^p$ are entries of estimated background $\mathbf{B}$ and
reference $\mathbf{R}$, respectively.

The proposed background initialization algorithm adopts illumination change
detection so it is robust when encountering gradual or sudden light changes.
SPMD considers both image spatial correlations and temporal motion detection. It
can remove relevant moving pixels more effectively. The density-based motionless
pixel clustering and final background determination generate reliable
background image $\mathbf{B}$ with low complexity. As a result, the proposed
algorithm is robust and effective for different situations and can generate a
clear background free of foreground objects for a wide range of video sequences.
Next section will present the detailed experimental results.

\section{Experimental Results}
\label{sec:result}

In this section, the performance of the proposed SPMD based background
initialization algorithm is evaluated in detail. Both subjective and objective
background performance are evaluated in Section~\ref{sec:performance}. The
results show that SPMD algorithm is comparable or superior to other
state-of-the-art works generating the best results in complex scenarios. In
addition, SPMD is a low-complexity algorithm. Section~\ref{sec:speed} demos that
it is at least twice faster than state-of-the-art works in terms of the
processing speed.

\subsection{Experimental Setup}

The experimental analysis is based on SBMnet
dataset\footnote{www.scenebackgroundmodeling.net}~\cite{Jodoin2017}. The dataset
includes 79 video sequences with resolution varying from 200$\times$144 to
800$\times$600. All sequences are grouped into 8 categories: \emph{Basic},
\emph{Intermittent Motion}, \emph{Clutter}, \emph{Jitter}, \emph{Illumination
	Changes}, \emph{Background Motion}, \emph{Very Long} and \emph{Very Short}. The
dataset includes a wide range of video sequences and can help to evaluate the
proposed algorithm objectively.

Six metrics are utilized to evaluate performance and they are briefly introduced.
To simplify the discussion, the ground truth and estimated background are
denoted as $\mathbf{B}_{GT}$ and $\mathbf{B}_E$, respectively.

\begin{enumerate}
	\item Average Gray-level Error (AGE): It is the average of the absolute
	difference between $\mathbf{B}_{GT}$ and $\mathbf{B}_E$. Images are converted to
	gray scale in advance.
	\item Percentage of Error Pixels (pEPs): An error pixel (EP) is the pixel whose
	value difference between $\mathbf{B}_{GT}$ and $\mathbf{B}_E$ is larger than a
	threshold $\tau$ ($\tau$ is set to be 20 in~\cite{Jodoin2017}). pEPs is defined
	as the ratio of EPs and total pixels. 
	\item Percentage of Clustered Error Pixels (pCEPs): A clustered error pixel
	(CEP) is an error pixel whose four connected pixels are also error pixels. pCEPs
	is defined as the ratio of CEPs and total pixels.
	\item Peak-Signal-to-Noise-Ratio (PSNR): PSNR is widely used to measure image
	quality. It is defined as 
	\begin{equation}
	PSNR=10 \cdot \log_{10}\biggl(\frac{MAX^2}{MSE}\biggr)
	\end{equation}
	$MSX$ is 255 and $MSE$ is the mean squared error between $\mathbf{B}_{GT}$ and
	$\mathbf{B}_E$.
	\item Multi-Scale Structural Similarity Index (MS-SSIM)~\cite{Wang2003}: This
	metric uses structural distortion (SSIM) as an estimation to compute perceptual
	image distortion, performing at multiple image scales.
	\item Color image Quality Measure (CQM)~\cite{Yalman2013}: CQM tries to measure
	perceptual image quality. PSNR values in YUV space are calculated and CQM is
	represented as 
	\begin{equation}
		CQM = PSNR_Y\times R_W + \frac{PSNR_U+PSNR_V}{2}\times C_W 
	\end{equation}
	$R_W$ and $C_W$ are two biologically-inspired coefficients set to 0.9449 and
	0.0551, respectively.
\end{enumerate}

The objective of background initialization is to minimize AGE, pEPs, pCEPs so
the background is accurate while PSNR, MS-SSIM and CQM should be maximized thus
the perceptual background quality is high. Next section will show the background
images estimated by our proposed SPMD method and the comparison with other
state-of-the-art works.

\begin{figure}[!t]
	\centering
	\setlength{\leftskip}{0pt plus 1fil minus \marginparwidth}
	\setlength{\rightskip}{\leftskip}
	\includegraphics[width = 5in]{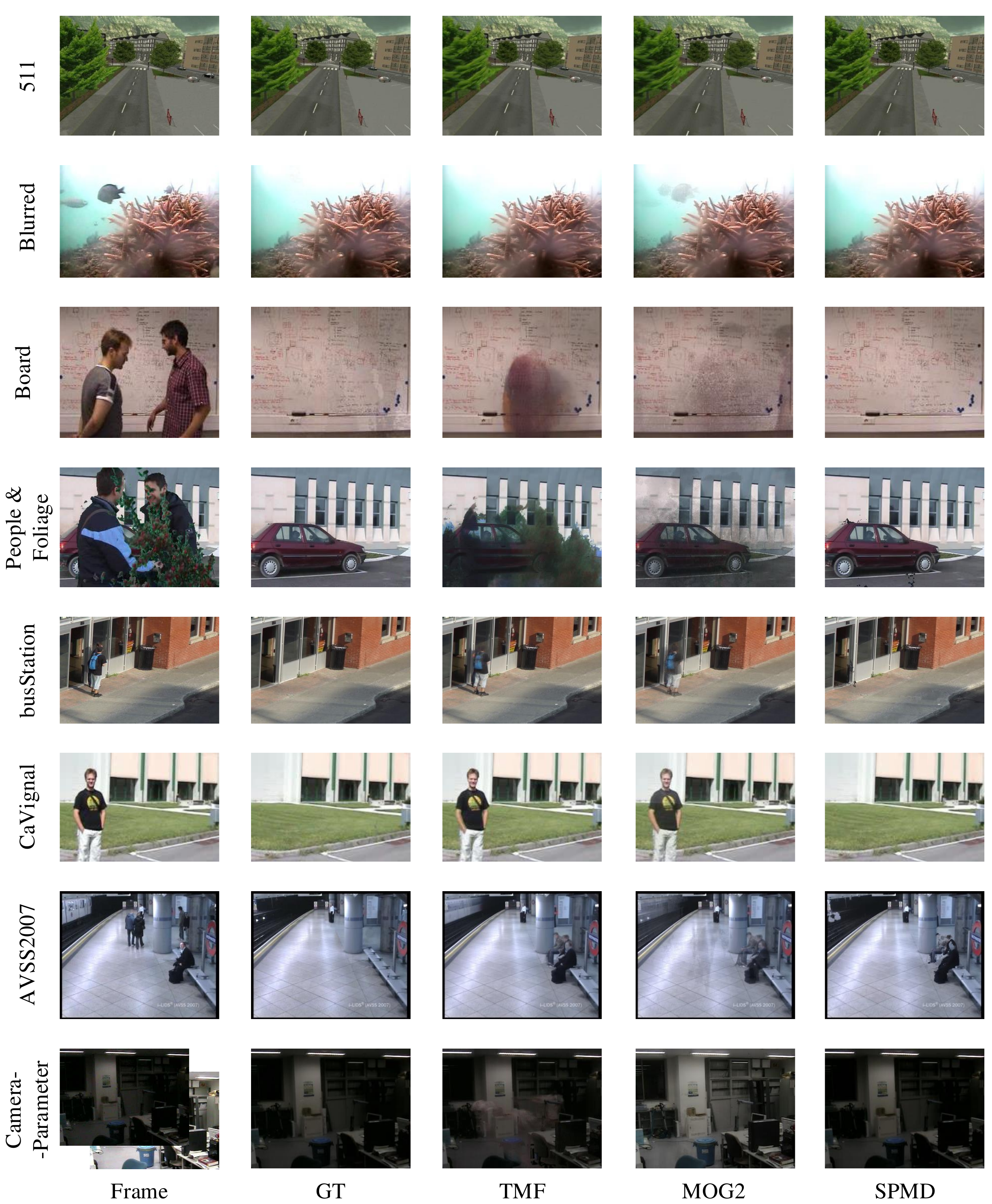}
	\caption{Visual background initialization results of the proposed algorithm for
		sequences of SBMnet dataset.}
	\label{fig:Result}
\end{figure}

\subsection{Evaluation Results and Comparison}
\label{sec:performance}

Fig~\ref{fig:Result} shows some examples of background initialization results of
the proposed algorithm. The first column is a representative frame in original
sequence, the second column is the true background image. Third column is the
results estimated by TMF and the fourth column is produced by MOG2. The results
of our proposed SPMD algorithm are shown in the last column. According to
Fig~\ref{fig:Result}, SPMD generages more clear backgrounds free of foreground
objects compared with TMF and MOG2.

The first two sequences, \emph{511} and \emph{Blurred} belongs to \emph{Basic}
category with mild object motion and long-term visible background, so all
methods initialize a clear background image with satisfied perceptual quality.
However, following sequences are more challenging. In \emph{Board}, the
background in the central area is less visible than foreground. In \emph{People
	\& Foliage}, the whole scene area is heavily occupied by foreground objects and
the background is divided into small isolated regions in each frame. TMF cannot
estimate correct background and results of MOG2 are also blurred by foreground
components, while SPMD successfully produces clear background images for these
two sequences. There exists intermittent object motion in following three
sequences \emph{busStation}, \emph{CaVignal} and \emph{AVSS2007}. For instance,
the man in \emph{CaVignal} stands still in the left of the scene for the first
60\% of sequence frames. Both TMF and MOG2 include foreground objects into the
estimated backgrounds and produce inaccurate image for these three sequences. On
the contrary, SPMD can remove motion objects more effectively and it estimates
an ideal background image for \emph{busStation} and \emph{CaVignal}. For
\emph{AVSS2007}, although SPMD does not remove all foreground objects, the
result is still better than other two methods. The last sequence
\emph{CameraParameter} exist strong illumination change caused by light
switching on. Because of the illumination change detection adopted in the
proposed algorithm, the initialized background is clear and accurate, which
shows the robustness of the proposed algorithm.

\begin{table}[!t]
	\caption{Overall Results Comparison of SPMD and Top 7 Methods on SBMnet.}
	\label{tab:Results}
	\centering
	\resizebox{\textwidth}{!}
	{
		\begin{tabular}{c|c|c|c|c|c|c|c}
			\hline
			\hline
			Method & AR & AGE & pEPs & pCEPs & MS-SSIM & PSNR & CQM \\
			\hline
			\hline
			SPMD                            & 1.83 & \textbf{6.0988} & \textbf{0.0488} & \textbf{0.0154} & \textbf{0.9412} & 29.8436 & 30.6827 \\ 
			\hline
			MSCL~\cite{Javed2017}           & 1.67 & \textbf{5.9547} & \textbf{0.0524} & \textbf{0.0171} & 0.9410 & \textbf{30.8952} & \textbf{31.7049} \\		
			LaBGen-OF~\cite{Laugraud2017_2} & 2.33 & 6.1897          & 0.0566          & 0.0232          & \textbf{0.9412} & \textbf{29.8957} & \textbf{30.7006} \\
			BEWiS~\cite{Gregorio2017}       & 4.67 & 6.7094          & 0.0592          & 0.0266          & 0.9282          & 28.7728 & 29.6342 \\
			LaBGen~\cite{Laugraud2017}      & 5.17 & 6.7090          & 0.0631          & 0.0265          & 0.9266          & 28.6396 & 29.4668 \\
			LaBGen-P~\cite{Laugraud2016}    & 6.33 & 7.0738          & 0.0706          & 0.0319          & 0.9278          & 28.4660 & 29.3196 \\
			Photomontage~\cite{Agarwala2004}& 6.50 & 7.1950          & 0.0686          & 0.0257          & 0.9189          & 28.0113 & 28.8719 \\
			SC\_SOBS-C4~\cite{Maddalena2016}& 7.33 & 7.5183          & 0.0711          & 0.0242          & 0.9160          & 27.6533 & 28.5601 \\
			\hline
			\hline
		\end{tabular}
	}
\end{table}

\begin{table}[!t]
	\caption{Evaluation Results Comparison of 8 Video Categories.}
	\label{tab:IndividualCategory}
	\centering
	\resizebox{\textwidth}{!}
	{
		\begin{tabular}{c|c|c|c|c|c|c|c}
			\hline
			\hline
			Category & Method & AGE & pEPs & pCEPs & MS-SSIM & PSNR & CQM \\
			\hline
			\hline
			\multirow{4}{*}{Basic} 
			& SPMD     & \textbf{3.8143} & 0.0119          & \textbf{0.0021} & \textbf{0.9812} & \textbf{33.9773} & \textbf{34.6012} \\ 
			& MSCL     & \textbf{3.4019} & \textbf{0.0112} & \textbf{0.0019} & \textbf{0.9807} & \textbf{35.1206} & \textbf{35.6507} \\ 
			& LaBGen-OF& 3.8421          & \textbf{0.0118} & 0.0033          & 0.9796          & 33.7714          & 34.3918 \\ 
			& BEWiS    & 4.0673          & 0.0162          & 0.0058          & 0.9770          & 32.2327          & 33.0035 \\ 
			\hline
			\multirow{4}{*}{Intermittent Motion} 
			& SPMD     & \textbf{4.1840} & \textbf{0.0207} & \textbf{0.0088} & \textbf{0.9745} & \textbf{31.9703} & \textbf{32.7384} \\ 
			& MSCL     & \textbf{3.9743} & 0.0313          & 0.0215          & \textbf{0.9831} & \textbf{32.6916} & \textbf{33.4541} \\ 
			& LaBGen-OF& 4.6433          & \textbf{0.0221} & \textbf{0.0120} & 0.9676          & 30.5799          & 31.3920 \\ 
			& BEWiS    & 4.7798          & 0.0277          & 0.0173          & 0.9585          & 29.7747          & 30.6778 \\ 
			\hline
			\multirow{4}{*}{Clutter} 
			& SPMD     & \textbf{4.6009} & \textbf{0.0250} & \textbf{0.0114} & 0.9572          & 30.9205          & 31.9057 \\ 
			& MSCL     & 5.2695          & 0.0275          & \textbf{0.0094} & \textbf{0.9629} & \textbf{31.3743} & \textbf{32.2837} \\ 
			& LaBGen-OF& \textbf{4.1821} & \textbf{0.0246} & 0.0117          & \textbf{0.9640} & \textbf{32.6339} & \textbf{33.4654} \\ 
			& BEWiS    & 10.6714         & 0.1227          & 0.0845          & 0.8610          & 25.4804          & 26.4783 \\ 
			\hline
			\multirow{4}{*}{Jitter} 
			& SPMD     & 9.8096          & 0.1123          & \textbf{0.0397} & 0.8501          & 24.3465          & 25.4892 \\ 
			& MSCL     & \textbf{9.7403} & \textbf{0.1049} & 0.0424          & 0.8475          & \textbf{25.3035} & \textbf{26.3824} \\ 
			& LaBGen-OF& \textbf{9.2410} & 0.1064          & \textbf{0.0380} & \textbf{0.8579} & \textbf{25.9053} & \textbf{26.9264} \\ 
			& BEWiS    & 9.4156          & \textbf{0.1048} & 0.0402          & \textbf{0.8524} & 24.7408          & 25.8579 \\ 
			\hline
			\multirow{4}{*}{Illumination Changes} 
			& SPMD     & \textbf{4.4752} & \textbf{0.0222} & \textbf{0.0090} & \textbf{0.9835} & \textbf{32.1318} & \textbf{32.9929} \\ 
			& MSCL     & \textbf{4.4319} & 0.0341          & \textbf{0.0134} & \textbf{0.9856} & \textbf{34.6735} & \textbf{35.3442} \\ 
			& LaBGen-OF& 8.2200          & 0.1130          & 0.0746          & 0.9654          & 27.8422          & 28.7690 \\ 
			& BEWiS    & 5.9048          & \textbf{0.0312} & 0.0223          & 0.9745          & 29.5427          & 30.3805 \\ 
			\hline
			\multirow{4}{*}{Background Motion} 
			& SPMD     & \textbf{9.9119} & \textbf{0.1252} & \textbf{0.0289} & \textbf{0.8587} & 25.5415          & 26.3356 \\ 
			& MSCL     & 11.2194         & 0.1540          & 0.0332          & 0.8448          & 24.4813          & 25.6982 \\ 
			& LaBGen-OF& 10.0698         & 0.1312          & 0.0323          & 0.8550          & \textbf{25.8626} & \textbf{26.6974} \\ 
			& BEWiS    & \textbf{9.6776} & \textbf{0.1258} & \textbf{0.0286} & \textbf{0.8644} & \textbf{26.0753} & \textbf{26.9685} \\ 
			\hline
			\multirow{4}{*}{Very Long} 
			& SPMD     & 6.0924          & 0.0283          & 0.0055          & 0.9824          & 30.3164          & 31.1255 \\ 
			& MSCL     & \textbf{3.8214} & 0.0172          & 0.0022          & 0.9874          & \textbf{32.2773} & \textbf{32.9941} \\ 
			& LaBGen-OF& 4.2856          & \textbf{0.0114} & \textbf{0.0006} & \textbf{0.9891} & 32.0746          & 32.8312 \\ 
			& BEWiS    & \textbf{3.9652} & \textbf{0.0108} & \textbf{0.0006} & \textbf{0.9891} & \textbf{32.5325} & \textbf{33.2217} \\ 
			\hline
			\multirow{4}{*}{Very Short} 
			& SPMD     & 5.9018          & 0.0446          & 0.0177          & 0.9423          & 29.5447          & 30.2731 \\ 
			& MSCL     & 5.7790          & 0.0387          & \textbf{0.0131} & 0.9363          & \textbf{31.2396} & \textbf{31.8320} \\ 
			& LaBGen-OF& \textbf{5.0338} & \textbf{0.0325} & \textbf{0.0135} & \textbf{0.9509} & \textbf{30.4959} & \textbf{31.1315} \\ 
			& BEWiS    & \textbf{5.1937} & \textbf{0.0345} & 0.0136          & \textbf{0.9488} & 29.8036          & 30.4858 \\ 
			\hline
			\hline
		\end{tabular}
	}
\end{table}

Besides subjective background image analysis, the results are quantitatively
compared with other state-of-the-art works. Top 7 methods on SBMnet dataset are
selected and they are listed here: MSCL~\cite{Javed2017},
LaBGen-OF~\cite{Laugraud2017_2}, BEWiS~\cite{Gregorio2017},
LaBGen~\cite{Laugraud2017}, LaBGen-P~\cite{Laugraud2016},
Photomontage~\cite{Agarwala2004} and  SC\_SOBS-C4~\cite{Maddalena2016}. The
overall results comparison is shown in Table~\ref{tab:Results}. AR means the
average ranking across 6 metrics of the total 8 algorithms and each metric is
calculated as the average of results on 79 sequences. Top performance is denoted
by bold numbers. Table~\ref{tab:Results} shows that the proposed SPMD achieves
the second best in AGE metric and the best in pEPs, pCEPs, MS-SSIM metrics.
Comparison results presents that SPMD outperforms LaBGen-OF, which is the
current second best method on SBMnet website. Our result is slightly worse than the
best method MSCL but some metrics are comparable. Overall, SPMD can estimate
accurate background for various sequences and the performance is comparable or
superior to other state-of-the-art works.

As SBMnet comprises 8 categories with different video characteristics, it is
necessary to evaluate performance of SPMD for individual category.
Table~\ref{tab:IndividualCategory} displays the evaluation results comparison
for 8 categories. Because of space limitation, SPMD is compared with the top 3
methods, MSCL, LaBGen-OF and BEWiS. For category \emph{Basic}, MSCL produces
the best results. SPMD achieves comparable performance in pCEPs and MS-SSIM metrics.
In other metrics, the performance is slightly worse than MSCL and ranks second.
For complex scenes in \emph{Intermittent Motion} and \emph{Clutter}, the
performance of SPMD is comparable with MSCL and LaBGen-OF, outperforming BEWiS.
In category \emph{Jitter}, all four algorithms produce similar results standing
for best performance. Because of the adoption of illumination change detection,
SPMD can estimate satisfied backgrounds similar to MSCL and the performance is
superior to other methods. For \emph{Background Motion}, SPMD has a close
performance to BEWiS, which is the best in this category. Although our algorithm
does not produces as good results as other three methods, the gap is very small
and the performance is acceptable. The comparison shows that SPMD is robust and
effective for a wide range of video sequences.

\subsection{Processing Speed Analysis}
\label{sec:speed}

\begin{table}[!t]
	\caption{Processing Speed Comparison for Sequences with Different Resolutions.}
	\label{tab:ProcessingTime}
	\centering
	{
		\begin{tabular}{c|c|c|c|c|c}
			\hline
			\hline
			Sequence   & Foliage & highway & I\_CA\_01 & wetSnow & 511 \\
			\hline
			Resolution & 200$\times$144 & 320$\times$240 & 352$\times$288 & 536$\times$320 & 640$\times$480 \\
			\hline
			SPMD & \textbf{22.8fps} & \textbf{5.6fps} & \textbf{5.2fps} & \textbf{2.9fps} & \textbf{1.6fps} \\
			LaBGen-OF & 3.3fps & 1.6fps & 1.2fps & 0.8fps & 0.5fps \\
			BEWiS & 4.3fps & 2.5fps & 2.2fps & 1.1fps & 0.6fps \\
			\hline
			\hline
		\end{tabular}
	}
\end{table}

Besides background image quality, the processing speed is also an important
aspect when evaluating the performance of algorithm. LaBGen-OF, BEWiS and SPMD
are all implemented on a computer with Intel core i7-2600 processor for
comparison. Five sequences with different resolutions from 200$\times$144 to
640$\times$480 are selected for evaluation. For a fair comparison, processing
frames per second (fps) is recorded to denote execution speed.
Table~\ref{tab:ProcessingTime} presents the performance in terms of processing
speed and it is noticed that SPMD is at least twice faster than other two
algorithms. Especially for \emph{Foliage}, the result of SPMD (22.8fps) is close
to a realtime computing speed. For LaBGen-OF, dense optical flow with high
computational complexity is performed to detect motion in a video sequence which
affects processing speed. In BEWiS, a weightless neural model is used for each
pixel with high computational and memory cost. For another state-of-the-art
algorithm MSCL, it is reported in~\cite{Javed2017} that the computing speed is
about 1.25fps for sequence \emph{highway} on a computer with Inter core i7
processor. The computational complexity of MSCL is also high because it adopts
dense optical flow for motion estimation and then an iterative optimization for
matrix completion is performed. Overall, the proposed SPMD algorithm has
relatively low computational complexity. The most time consuming period is
superpixel segmentation and it can be further accelerated with a CPU or GPU
based parallel implementation.

\section{Conclusions}
\label{sec:conclusion}

A background initialization algorithm based on superpixel motion detection
(SPMD) is proposed in this paper. A subsequence with stable illumination
condition is firstly selected making the result not affected by gradual or
sudden illumination changes. Images are segmented into superpixels so spatial
correlation is reserved. Frame difference based motion detection considering
temporal relation is combined with superpixel segmentation, thus SPMD can
effectively remove moving objects. Finally, the density-based clustering and the
background determination criteria automatically generate a clear background
image. Experimental results on SBMnet dataset show that the algorithm is robust
and effective for a wide range of videos with performance comparable or superior
with other existing works while the computational complexity is relatively low.

\section*{References}
\bibliography{reference}

\end{document}